\documentclass[sigconf]{acmart}
\AtBeginDocument{%
  }

\setcopyright{acmlicensed}
\copyrightyear{2018}
\acmYear{2018}
\acmDOI{XXXXXXX.XXXXXXX}
\acmConference[Conference acronym 'XX]{Make sure to enter the correct
  conference title from your rights confirmation email}{June 03--05,
  2018}{Woodstock, NY}
\acmISBN{978-1-4503-XXXX-X/2018/06}

\newcommand{\mlrdcn}{ML-DCN}

\begin{document}

%\title{Masked Lowrank DCN: Towards Scalable Pinterest Ads Ranking

\title{ML-DCN: Masked Low-Rank Deep Crossing Network Towards Scalable Ads Click-through Rate Prediction at Pinterest
}

\author{Jiacheng Li}
\email{jiachengli@pinterest.com}
%\email{webmaster@marysville-ohio.com}
\affiliation{%
  \institution{Pinterest}
  \country{USA}
}

\author{Yixiong Meng}
\email{ymeng@pinterest.com}
%\email{webmaster@marysville-ohio.com}
\affiliation{%
  \institution{Pinterest}
  \country{USA}
}

\author{Yi Wu}
\email{yiwu@pinterest.com}
%\email{webmaster@marysville-ohio.com}
\affiliation{%
  \institution{Pinterest}
  \country{USA}
}

\author{Yun Zhao}
\email{yunzhao@pinterest.com}
%\email{webmaster@marysville-ohio.com}
\affiliation{%
  \institution{Pinterest}
  \country{USA}
}

\author{Sharare Zehtabian}
\email{szehtabian@pinterest.com}
%\email{webmaster@marysville-ohio.com}
\affiliation{%
  \institution{Pinterest}
  \country{USA}
}

\author{Jiayin Jin}
\email{jjin@pinterest.com}
%\email{webmaster@marysville-ohio.com}
\affiliation{%
  \institution{Pinterest}
  \country{USA}
}

\author{Degao Peng}
\email{dpeng@pinterest.com}
%\email{webmaster@marysville-ohio.com}
\affiliation{%
  \institution{Pinterest}
  \country{USA}
}

\author{Jinfeng Zhuang}
\email{jzhuang@pinterest.com}
%\email{webmaster@marysville-ohio.com}
\affiliation{%
  \institution{Pinterest}
  \country{USA}
}

\author{Qifei Shen}
\email{qshen@pinterest.com}
%\email{webmaster@marysville-ohio.com}
\affiliation{%
  \institution{Pinterest}
  \country{USA}
}

\author{Kungang Li}
\email{kungangli@pinterest.com}
%\email{webmaster@marysville-ohio.com}
\affiliation{%
  \institution{Pinterest}
  \country{USA}
}

\renewcommand{\shortauthors}{Li et al.}

\begin{abstract}
%\jinfeng{make the title more specific. "ads Ranking" is too general - it can be retrieval, L1, etc. Probably have CTR prediction as the focus. Also think about how to make it more fancy: how a marriage of DCN and MaskNet is better than each single}

%\jinfeng{for tha authors, ask them to review/edit or contribute to the paper content. otherwise remove them from authors.}

%\jinfeng{make the motivation / logic in abstract clear \& clean: 1) we want scaling-up property; 2) we want efficiency; 3) we want a better model arch combining the benefit of DCN and MaskNet. So the reviewers will buy it at the first read.}

%\jinfeng{Add sth like "This novel combination of DCN and MaskNet, leads to a scaling-up property of the ranking model and produces state-of-the-art performance" - basically emphasize there are true innovations here}.

%\jinfeng{given it can be launched in RP combo, mention clearly that we have deployed this into production.}

Deep learning recommendation systems rely on feature interaction modules to model complex user–item relationships across sparse categorical and dense features. In large-scale ad ranking, increasing model capacity is a promising path to improving both predictive performance and business outcomes, yet production serving budgets impose strict constraints on latency and FLOPs. This creates a central tension: we want interaction modules that both scale effectively with additional compute and remain compute-efficient at serving time. In this work, we study how to scale feature interaction modules under a fixed serving budget. We find that naively scaling DCNv2 and MaskNet, despite their widespread adoption in industry, yields rapidly diminishing offline gains in the Pinterest ads ranking system. To overcome aforementioned limitations, we propose \mlrdcn, an interaction module that integrates an instance-conditioned mask into a low-rank crossing layer, enabling per-example selection and amplification of salient interaction directions while maintaining efficient computation. This novel architecture combines the strengths of DCNv2 and MaskNet, scales efficiently with increased compute, and achieves state-of-the-art performance. Experiments on a large internal Pinterest ads dataset show that \mlrdcn\ achieves higher AUC than DCNv2, MaskNet, and recent scaling-oriented alternatives at matched FLOPs, and it scales more favorably overall as compute increases, exhibiting a stronger AUC–FLOPs trade-off. Finally, online A/B tests demonstrate statistically significant improvements in key ads metrics (including CTR and click-quality measures) and \mlrdcn \space has been deployed in the production system  with neutral serving cost.

\end{abstract}





\maketitle

\section{Introduction}
Modern recommendation systems based on deep learning models constitute the core components of a wide range of online platforms and digital services (\cite{covington2016deep,ying2018graph,pal2020pinnersage,badrinath2025omnisage,zhuang2025practice,zhang2022dhen,zhang2021deep}). At a high level, advanced deep learning recommendation systems work by analyzing a mixture of signals from multiple sources. Data describing users, items, and user--item interactions is typically transformed into dense (numerical) features (e.g., age) and sparse categorical features (e.g., an advertiser ID). The recommendation system maps each sparse categorical feature to an embedding vector using a learnable dictionary-like structure called an embedding table. These dense representations are then fed into a dedicated interaction module, which is designed to model the intricate interactions between features.

In large-scale ads ranking systems, model scaling is constrained by strict production requirements, including inference latency, throughput, and hardware cost. A common scaling approach is to expand embedding tables by increasing the number of IDs and embedding dimensions, which can reduce hash collisions and improve expressiveness (\cite{su2025multi,kang2021learning,lian2022persia,mudigere2021high}). More specifically, our recent work on scaling up embedding tables and leveraging a distributed embedding-table training paradigm~\cite{su2025multi} achieved significant gains on top-line ads metrics. However, this scaling direction is increasingly memory-capacity and memory-bandwidth bound and does not fully utilize modern accelerators, since larger embedding tables primarily increase memory requirements rather than proportionally increasing compute utilization. Moreover, as embedding memory grows, the marginal gains can diminish while infrastructure cost continues to rise, motivating alternative scaling mechanisms that increase model expressiveness with a more favorable compute--cost profile.

The goal of this work is to scale recommendation models by increasing interaction-module capacity while keeping overall compute and serving cost under control. Popular interaction architectures such as DCNv2 \cite{wang2021dcn} and MaskNet \cite{wang2021masknet} are already deployed in Pinterest ads recommendation. However, in our production feature set and training setup, we find that naively scaling DCNv2- and MaskNet-based interaction modules yields rapidly diminishing offline AUC gains, indicating saturation in the compute regimes we can afford.

To address this, we propose \mlrdcn \space (Masked Lowrank DCN), a new interaction unit for large-scale CTR prediction. \mlrdcn \space builds on the low-rank DCN formulation for compute efficiency and introduces an instance-guided masking mechanism \emph{inside the low-rank interaction space}, enabling per-example selection and amplification of salient interaction directions under a constrained FLOPs budget. In offline experiments on a large internal Pinterest ads dataset, \mlrdcn \space achieves higher AUC at matched FLOPs compared with DCNv2 and MaskNet, i.e., an improved AUC--FLOPs trade-off. We further compare our approach with recent methods such as WuKong \cite{zhang2024wukong} and RankMixer \cite{zhu2025rankmixer}, which report promising scaling trends relating AUC improvements to dense parameter count and/or FLOPs. \mlrdcn \space exhibits a more favorable AUC--FLOPs scaling trend than these recent approaches in our setting, suggesting that it is particularly effective at modeling high-order user--item feature interactions. Finally, we conducted online A/B tests to validate that offline gains translate to online impact and observed statistically significant improvements in key advertising metrics, including a $+1.89\%$ relative increase in overall ads click-through rate (CTR), while keeping serving cost neutral relative to the production baseline.

The key contributions of this paper are:
\begin{itemize}
\item A new interaction module, Masked Lowrank DCN. We introduce a low‑rank cross network augmented with an instance guided masking mechanism that selectively amplifies salient feature interaction directions while keeping FLOPs modest.
\item Scalable interaction capacity under serving constraints. We show that Masked Lowrank DCN achieves a more favorable AUC–FLOPs trade‑off than DCNv2, MaskNet, WuKong, and RankMixer on a large‑scale Pinterest ads CTR prediction task, especially in the higher‑compute regimes where existing interaction modules saturate.
\item Comprehensive empirical and production validation. Through ablation experiments, we quantify the impact of the number of layers, internal rank, mask ratio, and mask components, demonstrate that Masked Lowrank DCN generalizes well as an expert in an MMoE architecture, and validate that the offline improvements translate to statistically significant lifts in CTR and click‑quality metrics in large‑scale online A/B tests at neutral serving cost relative to the production baseline.
\end{itemize}
\section{Related Work} 
A common paradigm for recommendation models is to use hybrid architectures that learn explicit and implicit feature interactions in parallel. The foundational DeepFM~\cite{guo2017deepfm} model popularized this design by combining a Factorization Machine (FM) for efficiently capturing low-order explicit feature crosses with a standard multi-layer perceptron (MLP) that learns high-order implicit interactions. Following this pattern, xDeepFM~\cite{lian2018xdeepfm} extended the idea by replacing the FM component with a more powerful Compressed Interaction Network (CIN). The CIN was designed to generate explicit feature interactions in a vector-wise fashion, providing a more expressive way to model explicit feature crosses than the original FM.

Beyond these, researchers have introduced a variety of other interaction architectures. The Deep \& Cross Network (DCN)~\cite{wang2017deep} and its successor, DCNv2~\cite{wang2021dcn}, generate explicit feature crosses with increasing polynomial degree via a cross network (typically in a vector-wise form). In contrast, MaskNet~\cite{wang2021masknet} applies an instance-guided mask to MLP hidden representations, using element-wise gating to modulate feature interactions. FiBiNET~\cite{huang2019fibinet} uses a Squeeze-and-Excitation network to reweight feature importance before modeling interactions, while the Adaptive Factorization Network~\cite{cheng2020adaptive} (AFN) was designed to adaptively learn the order of feature crosses. Moreover, FinalMLP~\cite{su2023embarrassingly} shows that improvements can be obtained by improving the final fusion stage. Finally, AutoInt~\cite{song2019autoint} leverages attention mechanisms to learn feature relationships automatically.

Recent advancements such as WuKong~\cite{zhang2024wukong} and RankMixer~\cite{zhu2025rankmixer} suggest that the interaction module can exhibit favorable scaling behavior reminiscent of scaling trends observed in LLMs. Historically, recommendation models have not always improved predictably with scale due to inefficient scaling mechanisms and architectures that do not fully utilize modern accelerators. WuKong addresses this by proposing a simple architecture composed of stacked Factorization Machine (FM) blocks, enabling it to capture diverse high-order feature interactions. The authors report consistent scaling trends over roughly two orders of magnitude in model size on large-scale datasets. Similarly, RankMixer introduces a hardware-aware unified architecture designed to maximize GPU utilization. It replaces expensive self-attention with a lightweight multi-head token-mixing module and uses per-token Feed-Forward Networks (FFNs), which can be expanded with a Sparse Mixture-of-Experts (SMoE) to substantially increase model capacity with limited additional compute. By decoupling parameter growth from computational cost, RankMixer is scaled up by 100$\times$ while maintaining inference latency, suggesting that with domain-specific, hardware-aware designs, recommendation models can benefit substantially from scaling.

Ensemble-style designs have also been explored for recommendation, where model capacity is increased by combining multiple specialized components. Multi-gate Mixture-of-Experts (MMoE)~\cite{ma2018modeling} learns a shared set of experts and uses task-specific gates to dynamically combine expert outputs, enabling selective parameter sharing and improved modeling flexibility. In a complementary direction, Deep and Hierarchical Ensemble Network (DHEN)~\cite{zhang2022dhen,zhuang2025practice} hierarchically ensembles heterogeneous interaction modules, leveraging the complementary inductive biases of different interaction operators to strengthen overall representation power.

%\jinfeng{This is pitched as a single corssing module. Probably mention there are enselbme networks (DHEN e.g.), and ML-DCN can be combined into ensembles in the future.}
\section{Proposed Architecture: \mlrdcn}
\subsection{Background: Ads Ranking Models in Pinterest}
The core of Pinterest's advertising ecosystem is a sophisticated recommendation engine designed to deliver highly relevant and personalized ad content to a global audience of hundreds of millions. The effectiveness of this platform hinges on the predictive accuracy of its ranking algorithms, which are primarily tasked with estimating user engagement. This is quantified through key metrics such as the click-through rate (CTR) and the conversion rate (CVR). In this paper, we mainly discuss the CTR prediction model. The CTR, a fundamental measure of engagement, is defined as the probability that a user will click on an advertisement after it is shown:

\begin{equation*}
    CTR = P(\text{click}\mid \text{impression}).
\end{equation*}

Modern ads ranking models are mostly based on deep learning, which are capable of consuming abundant information from thousands of input features and making accurate predictions ~\cite{li2023deep, liu2022review}. At a high level, Pinterest ads ranking models utilize three types of features, including 
\begin{itemize}
\item Pin(item) features denoted by $P = \big\{P_1, P_2, ..., P_{N_1} \big\}$, such as Prior CTR of a Pin aggregated over some time window, Pin ID, advertiser ID and content embeddings, etc. 
\item User features denoted by $U = \big\{U_1, U_2, ..., U_{N_2}\big\}$, including counting features such as available user demographic features, user embeddings and sequence features built based on users' past interaction history. 
\item User-Pin interaction features denoted by $I = \big\{I_1, I_2, ..., I_{N_3} \big\}$. For example, user-advertiser interaction counts and user engagement history based features, etc.  
\end{itemize}

Given a user and a Pin, our ads ranking system fetches the relevant features and inputs them into a CTR model. The CTR prediction model can be represented as a function $$f_{CTR}(P, U, I) \rightarrow [0, 1]$$.

\subsection{Review of DCNv2 and MaskNet}
In recommendation systems, input features are typically grouped into two categories: dense (numerical) features and categorical features. Dense features are commonly preprocessed using normalization, logarithmic transformations, or more advanced mixture feature transformations~\cite{zhuang2020feature}. Categorical features are mapped to learned representations by looking up their corresponding embeddings from embedding tables. Following the notation from DCNv2~\cite{wang2021dcn}, we use $x_0 \in \mathbb{R}^{d}$ to denote the input features to the interaction modules. Notice that the $x_0$ is already a combination of the embeddings of categorical features and processed dense features. We use $X_0 \in \mathbb{R}^{B \times d}$ to denote a batch of input features, where $B$ is the batch size. 

\subsubsection{DCNv2}
The core of DCNv2 is the cross layers that create explicit features crosses, as shown in Eq.(\ref{eq:dcnv2}) for the $(l+1)^{\text{th}}$ cross layer:
\begin{equation}
    X_{l+1} = X_0 \odot \left(X_l W_l + b_l\right) + X_l
    \label{eq:dcnv2}
\end{equation}

where $X_0 \in \mathbb{R}^{B \times d}$ is input to the first crossing layer. $X_l$ and $X_{l+1}$ are the input and the output of $(l+1)^{\text{th}}$ crossing layer. $W_l \in \mathbb{R}^{d \times d}$ and $b_l \in \mathbb{R}^d$ are the learned weight matrix and bias vector. The highest polynomial order is $l+1$ for an $l$-layered cross network.

On top of the DCNv2 cross layer formulation, a more cross-effective low-rank DCN is proposed in the same research paper. For each low-rank cross layer, the formulation is shown in Eq.(\ref{eq:lowrankdcnv2}):
\begin{equation}
    X_{l+1} = X_0 \odot \left(\left(X_l V_l\right)U_l^{\top} + b_l\right) + X_l
    \label{eq:lowrankdcnv2}
\end{equation}

where $U_l,V_l \in \mathbb{R}^{d \times r}$ and $r \ll d$. 

Furthermore, this low-rank DCN structure can be extended to the Mixture-of-Experts(MoE) format with a gating mechanism ~\cite{eigen2013learning,shazeer2017outrageously,ma2018modeling,jacobs1991adaptive}. 
The resulting mixture of low-rank cross layer formulation is shown in Eq. (\ref{eq:moe_update}) and Eq. (\ref{eq:moe_expert}):
\begin{align}
    X_{l+1} &= X_l + \sum_{i=1}^{K} G_i(X_l)\odot E_i(X_l)
    \label{eq:moe_update}
\end{align}

\begin{align}
    E_i(X_l) &= X_0 \odot \left(\left(X_l V_l^i\right)\left(U_l^i\right)^{\top} + b_l\right)
    \label{eq:moe_expert}
\end{align}

where $K$ is the number of experts and the gating network outputs
$G(X_l)\in\mathbb{R}^{B\times K}$.
The $i$-th gate $G_i(X_l)\in\mathbb{R}^{B\times 1}$ assigns a scalar weight
to expert $i$ for each instance in the batch (broadcast over the $d$ feature dimensions).
Typically, $G(X_l)$ is normalized with a softmax over experts for each instance.
When $K=1$ and $G_1(X_l)\equiv \mathbf{1}_{B\times 1}$, Eq (3) falls back to Eq (2).

%A natural extension is to increase the expressiveness of the cross layer by adding a non‑linear transformation in the low‑rank projection space, as shown in the Eq. (\ref{eq:moe_comp}):
%\begin{equation}
%    E_i(X_l) = X_0 \odot \left(
%        \left(g\!\left(g\!\left(X_l V_l^i\right) %C_l^i\right)\right)\left(U_l^i\right)^{\top} + b_l
%    \right)
%    \label{eq:moe_comp}
%\end{equation}

%where $g(\cdot)$ represents any nonlinear activation function, like tanh or ReLU~\cite{lecun1991efficient,nair2010rectified}, and $C^i_l \in \mathbb{R}^{r \times r}$ is a learnable matrix. 

\subsubsection{MaskNet}
Following the same notation from the previous subsection, here we keep using $X_0$ as the batched input to the interaction modules, where each row is a concatenated combination of sparse features and dense features. For the $(l+1)^{\text{th}}$ MaskBlock, the input is $X_{l}$ and the output is $X_{l+1}$. There are three key steps for each MaskBlock: instance-guided mask generation, masked embedding generation and final output computation~\cite{wang2021masknet}.

The instance-guided mask $X_{mask}$ is generated as:
\begin{equation}
    X_{mask} = \mathrm{ReLU}\!\left(X_l W_l^{d1} + b_l^{d1}\right) W_l^{d2} + b_l^{d2}
    \label{eq:mask}
\end{equation}
where $W_l^{d1} \in \mathbb{R}^{d \times k}$ and $W_l^{d2} \in \mathbb{R}^{k \times d}$ are weight matrices of the aggregation and projection layers, respectively. The dimension $k$ is typically larger than $d$ to form a bottleneck structure and the ratio $t=\frac{k}{d}$ is referred as the mask ratio. 

With the instance-guided mask, the masked embedding $X_{masked}$ is computed as:
\begin{equation}
    X_{masked} = X_{mask} \odot \mathrm{LN}(X_l)
\end{equation}
where $\mathrm{LN}(\cdot)$ denotes LayerNorm~\cite{ba2016layer}.
Then, the final output is computed as:
\begin{equation}
    X_{l+1} = \mathrm{ReLU}\!\left(\mathrm{LN}\!\left(X_{masked} W_l^{d3}\right)\right)
\end{equation}
where $W_l^{d3} \in \mathbb{R}^{d \times d}$ is a learnable weight matrix. The paper proposes two ways to use these blocks: in series (Stack) or in parallel.

%\jinfeng{Use bullet point to show the insights and importance of this combination, this is the key to convince reviewers why this is a good paper.}
\subsection{\mlrdcn} 
In an ideal setting, the interaction module should scale to larger model capacities while keeping infrastructure costs modest. Prior work~\cite{zhu2025rankmixer,zhang2024wukong} suggests that DCNv2 and MaskNet do not scale effectively under increasing compute, including low-rank DCNv2 variants. To address this limitation, we propose \mlrdcn, which delivers substantial offline improvements when scaled, while retaining a low-rank structure that keeps serving cost manageable.

Our design builds on two complementary components:
\begin{itemize}
    \item \textbf{Low-rank DCNv2 backbone for efficiency.} We adopt a low-rank variant of DCNv2 as the core crossing module, which parameterizes feature crossing in a low-rank form and thereby reduces parameters and FLOPs, leading to lower serving cost.
    \item \textbf{Instance-guided masking for effectiveness.} To compensate for reduced capacity in the low-rank regime, we introduce an instance-conditioned mask that also operates in a low-rank form to dynamically select and amplify salient interaction directions on a per-example basis, while suppressing less informative interactions. This targeted allocation of interaction capacity improves predictive performance under a comparatively low FLOPs budget.
\end{itemize}

%Our approach builds on low-rank DCNv2 to control computation, and improves expressiveness by prioritizing the most informative feature interactions. In low-rank regimes, surpassing full-rank DCNv2 requires selecting salient interactions while suppressing less relevant ones. Accordingly, we introduce an instance-guided masking mechanism as a key complement to the low-rank crossing module, enabling per-example selection and amplification of important interaction directions. By allocating additional interaction capacity only to high-importance features, \mlrdcn\ improves performance under a comparatively low FLOPs budget.

\begin{figure}
    \centering
    \includegraphics[width=1\linewidth,height=1.5\linewidth]{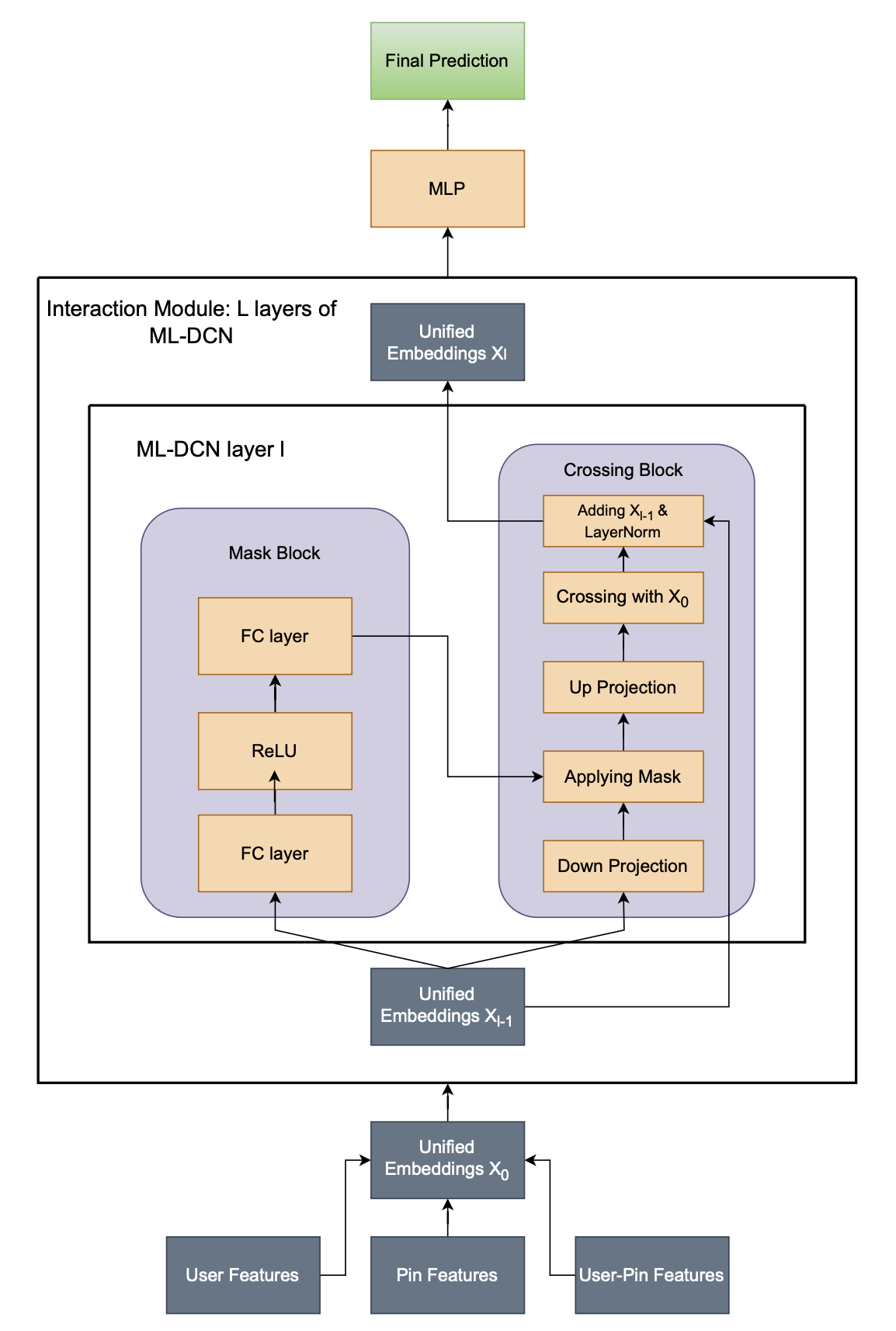}
    \caption{Architecture of a \mlrdcn \space block. Given input $X_{l-1}$, the block projects inputs into a low-rank interaction space, applies an instance-guided mask computed from $X_{l-1}$, maps the masked interactions back to the original input space and crosses with $X_0$, then adds a residual connection and LayerNorm to produce $X_l$.}
    \label{fig:mlrdcn}
\end{figure}

The detailed architecture of \mlrdcn \space is illustrated in Figure \ref{fig:mlrdcn}. Following the same notation as previous subsections, for the $(l+1)^{\text{th}}$ \mlrdcn \space block, the input is $X_{l}$ and the output is $X_{l+1}$. The output of the $(l+1)^{\text{th}}$ layer is computed as: 
\begin{equation}
    X_{l+1} =
    \mathrm{LN}\!\left(
        X_0 \odot
        \left(
            \left(\left(X_l V_l\right) \odot X_{mask}\right) U_l^{\top} + b_l
        \right)
        + X_l
    \right)
    \label{eq:condense_mlrdcn}
\end{equation}

where $U_l,V_l \in \mathbb{R}^{d \times r}$, $r \ll d$, $X_{mask} \in \mathbb{R}^{B \times r}$ and $\mathrm{LN}(\cdot)$ denotes LayerNorm. In addition, the instance-guided mask $X_{mask}$ is computed in the same way as Eq. (\ref{eq:mask}).

From a scaling perspective, DCNv2 exhibits a limited growth in interaction order: with $l$ cross layers, its highest polynomial order is $l+1$. In contrast, \mlrdcn\ increases the maximum interaction order substantially faster as depth grows, due to the instance-guided masking mechanism. Specifically, with $l$ layers, the highest polynomial order of \mlrdcn\ is $2^{l+1}-1$. This suggests that \mlrdcn\ can represent higher-order feature interactions more effectively than DCNv2 and MaskNet, while still maintaining a favorable infrastructure footprint through its low-rank parameterization. Taken together—low computational cost and stronger expressivity—\mlrdcn\ is a natural candidate for scaling-law exploration in ads recommendation systems, and our results on Pinterest internal data indeed demonstrate more favorable scaling behavior than the compared alternatives.

\section{Experiments and Results}
\subsection{Offline Experiments}
We conduct offline experiments for ads CTR prediction using an internal Pinterest dataset. We compare our proposed \mlrdcn\ against four alternative architectures: DCNv2, MaskNet, WuKong, and RankMixer. DCNv2 and MaskNet serve as strong baseline interaction modules, while WuKong and RankMixer represent recent advances that report favorable scaling behavior. Unless otherwise stated, all reported AUC gains are measured relative to the production baseline model, and FLOPs are computed for the \emph{entire} model (per inference) rather than the interaction module alone. We treat an AUC gain of $0.1\%$ as practically significant in our production setting. For brevity, we omit additional architectures such as AutoInt~\cite{song2019autoint} and FinalMLP~\cite{su2023embarrassingly}, as preliminary experiments in our production setting did not yield consistent AUC improvements.

\begin{figure}
    \centering
    \includegraphics[width=1\linewidth]{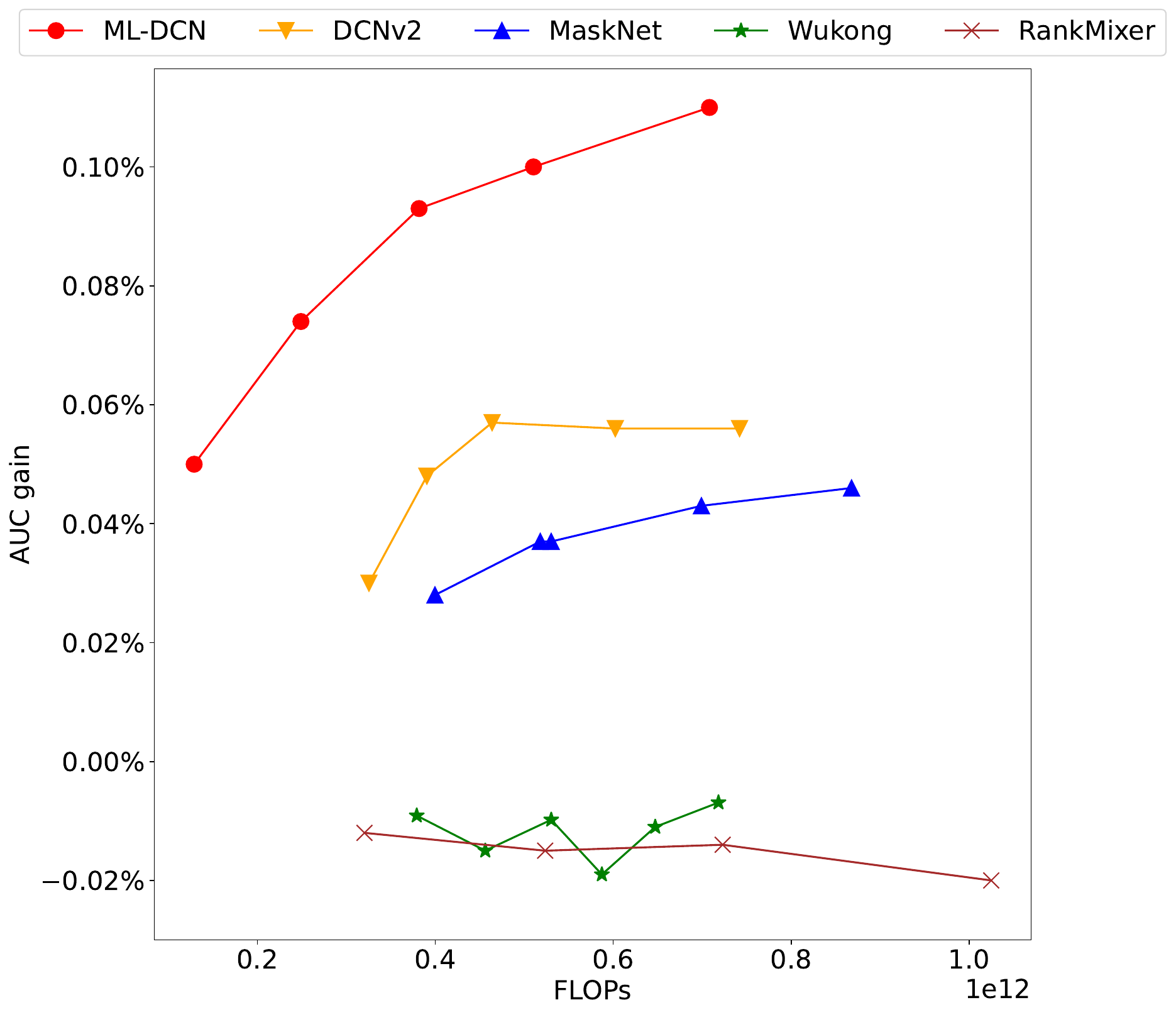}
    \caption{The AUC gain versus FLOPs curve.}
    \label{fig:scaling_law}
\end{figure}

\textbf{Scaling results.} Figure~\ref{fig:scaling_law} summarizes the scaling behavior of the five architectures by plotting AUC gain versus FLOPs. The proposed \mlrdcn\ consistently outperforms all other architectures at matched FLOPs, demonstrating its compute efficiency. Moreover, \mlrdcn\ achieves the best AUC--FLOPs trade-off among all evaluated models, including MaskNet and DCNv2, which further supports its ability to scale effectively with increased compute.

For DCNv2, we observe clear saturation: beyond a moderate compute regime, additional FLOPs yield diminishing returns and the AUC gain plateaus. This aligns with prior observations that DCNv2 often benefits most from 3--4 cross layers, with deeper stacks providing limited additional improvement~\cite{zhang2024wukong}. MaskNet improves as FLOPs increase, but remains less FLOPs-efficient than \mlrdcn, which matches our motivation of exploiting low-rank structure for efficient feature interactions. In contrast, RankMixer and WuKong do not yield measurable AUC improvements under our production feature set and training setup, and scaling their compute budgets does not translate into additional gains.

\subsubsection{Ablation study}
In this ablation study, we study the impact of key \mlrdcn\ hyperparameters: the number of \mlrdcn\ layers $l$, the rank $r$ of the low-rank projection matrices $U$ and $V$, and the mask ratio $t$. We also study the effect of the masking mechanism and several mask variants. We conclude with an MMoE experiment, in which \mlrdcn\ serves as the expert module.

\textbf{Impact of number of layers $l$.}
In Table~\ref{tab:num_layers}, we present an ablation on the number of layers for \mlrdcn, while fixing the rank $r$ at $4096$ and the mask ratio $t$ at $0.5$. We observe that increasing the number of layers yields larger AUC gain.

\textbf{Impact of internal dimension $r$.}
In Table~\ref{tab:internal_dim}, we present an ablation on the internal dimension $r$ for \mlrdcn, while fixing the number of layers $l$ at $4$ and the mask ratio $t$ at $0.5$. We observe that increasing $r$ yields larger AUC gain, consistent with increased model capacity.

\begin{table}[]
    \centering
    \begin{tabular}{c|c}
    \toprule
        number of layers $l$ & AUC gain  \\
        \midrule
        2 & 6.5e-4 \\
        3 & 8.1e-4 \\
        4 & 9.2e-4 \\
        5 & 9.7e-4 \\
        6 & 9.9e-4 \\
    \bottomrule
    \end{tabular}
    \caption{The impact of number of layers on AUC gain.}
    \label{tab:num_layers}
\end{table}

\begin{table}[]
    \centering
    \begin{tabular}{c|c}
    \toprule
        internal dimension $r$ & AUC gain  \\
        \midrule
        2048 & 7.5e-4 \\
        3072 & 8.2e-4 \\
        4096 & 9.3e-4 \\
        5120 & 9.8e-4 \\
        6144 & 1.0e-3 \\
    \bottomrule
    \end{tabular}
    \caption{The impact of internal dimension on AUC gain.}
    \label{tab:internal_dim}
\end{table}

\textbf{Impact of mask ratio $t$.}
In this experiment, we evaluate four mask ratios, $t \in \{0.5, 1, 1.5, 2\}$, while fixing the internal dimension to $r=4096$ and the number of layers to $l=4$. The results are reported in Table~\ref{tab:reduction_ratio}. For our use case, we find that a relatively small mask ratio (e.g., $t=0.5$) preserves most of the AUC improvement. This differs from the default choice $t=2$ in MaskNet~\cite{wang2021masknet}, which is motivated by using an ``aggregation'' layer (the first fully connected layer) that is wider than the subsequent ``projection'' layer.

We hypothesize that the discrepancy arises from the low-rank formulation adopted in \mlrdcn. Because the low-rank constraint already biases the model toward a small set of salient interaction patterns, the mask does not require a highly over-parameterized aggregation layer to capture useful signals. Moreover, making the aggregation layer narrower than the projection layer can act as an additional bottleneck, strengthening feature selection and better aligning with the capacity-controlled nature of the low-rank setting.

\begin{table}[]
    \centering
    \begin{tabular}{c|c}
    \toprule
        Mask ratio $t$ & AUC gain  \\
        \midrule
        0.5 & 9.2e-4 \\
        1 & 9.4e-4 \\
        1.5 & 9.7e-4 \\
        2 & 1.0e-3 \\
    \bottomrule
    \end{tabular}
    \caption{The impact of mask ratio on AUC gain.}
    \label{tab:reduction_ratio}
\end{table}

\textbf{Impact of each component of the mask.}
In this experiment, we fix the \mlrdcn\ configuration to internal dimension $r=4096$, $l=4$ layers, and mask ratio $t=0.5$, and conduct an ablation study on the masking mechanism. Specifically, we isolate the contribution of each mask component---(i) the first MLP layer, (ii) the ReLU nonlinearity, and (iii) the second MLP layer---by progressively removing them. The results are summarized in Table~\ref{tab:mask_ablation}. When the masking mechanism is removed, the model reduces to a low-rank DCNv2 cross network with fixed internal dimension $r=4096$ and $l=4$ layers; we denote this setting as ``No Mask''. Overall, incorporating additional mask components yields consistent AUC improvements, indicating that each element of the masking mechanism contributes meaningfully to model effectiveness.

\begin{table}[]
    \centering
    \begin{tabular}{c|c}
    \toprule
        Setting & AUC gain \\
        \midrule
        No Mask & 3.0e-4 \\
        1st MLP  & 6.8e-4 \\
        1st MLP + ReLU  & 8.1e-4 \\
        1st MLP + ReLU + 2nd MLP & 9.2e-4 \\
    \bottomrule
    \end{tabular}
    \caption{Ablation study for each component of the mask.}
    \label{tab:mask_ablation}
\end{table}

\textbf{Extension on MMoE.}
This experiment investigates the compatibility of \mlrdcn\ with the MMoE architecture~\cite{jacobs1991adaptive,ma2018modeling}. Each \mlrdcn\ expert shares the same configuration: internal dimension $r=4096$, $l=4$ layers, and mask ratio $t=0.5$. We sweep the number of experts in the MMoE block---used as the primary feature-interaction module within the recommendation model---by progressively adding experts. Due to computational resource constraints, we evaluate up to three experts; the corresponding results are summarized in Table~\ref{tab:mmoe}. Increasing the number of experts yields monotonic AUC improvements, indicating that additional expert capacity is beneficial in this setting.

\begin{table}[]
    \centering
    \begin{tabular}{c|c}
    \toprule
        number of experts & AUC gain  \\
        \midrule
        1 & 9.2e-4 \\
        2 & 1.1e-3 \\
        3 & 1.2e-3 \\
    \bottomrule
    \end{tabular}
    \caption{Experimental results for an MMoE structure with \mlrdcn\ as the expert module.}
    \label{tab:mmoe}
\end{table}

\subsection{Online Experiments}
We conduct an online A/B test to validate whether offline gains translate to online performance, focusing on ads click-through rate (CTR) prediction. In this experiment, we use \mlrdcn\ as the primary feature-interaction module, with internal dimension $r=4096$, $l=4$ layers, and mask ratio $t=0.5$. Table~\ref{tab:online_results} reports the resulting online gains. In addition to standard business metrics---cost per click (CPC) and click-through rate (CTR)---we define two CTR-quality metrics to better assess click quality in the online setting:
\begin{itemize}
    \item \textbf{Good CTR (gCTR):} The proportion of clicks where the session duration exceeds 30 seconds.
    \item \textbf{Outbound CTR (oCTR):} The proportion of clicks that lead to external websites or landing pages.
\end{itemize}

\begin{table}[h]
    \centering
    \begin{tabular}{l c}
        \toprule
        \textbf{Online Metric} & \textbf{Relative change} \\
        \midrule
        Ads CPC & -0.93\%\\
        Platform-wide CTR & +1.89\%\\
        Platform-wide gCTR & +2.17\%\\
        Platform-wide oCTR & +1.90\%\\
        \bottomrule
    \end{tabular}
    \caption{Online performance of \mlrdcn\ for CTR prediction.}
    \label{tab:online_results}
\end{table}

We observe improvements across all core online metrics, while infrastructure cost remains neutral compared to the existing production model.
\section{Conclusion and Future Work}

In summary, our work proposes a novel interaction module, \mlrdcn. Through extensive offline experiments, we show that \mlrdcn\ consistently improves AUC as model capacity scales. \mlrdcn\ also outperforms existing interaction modules at matched FLOPs. Moreover, the flexibility provided by the tunable internal dimension $r$, number of layers $l$, and mask ratio $t$ can be leveraged to customize \mlrdcn\ to a range of real-world applications with different data volumes and feature compositions. Finally, online A/B tests validate the offline gains, showing meaningful lifts in key ads metrics, including $+1.89\%$ relative increase in platform-wide CTR.

For future work, an important next step is to scale \mlrdcn\ further with larger compute budgets and more powerful computational resources. We also plan to investigate more expressive feature interaction mechanisms to further improve performance. Finally, we will focus on systems-level optimization of \mlrdcn\ to reduce training and inference latency and improve overall efficiency.

\section{Acknowledgment}
We would like to express our sincere gratitude to all those who contributed to the development of our online ads recommendation model. Our thanks also go to Abhijit Pol for their general support. Their expertise, dedication,  collaboration and support were essential to the success of this project.


%\bibliographystyle{ACM-Reference-Format}
%\bibliography{sample-base}
%\input{appendix}
\end{document}